\documentclass[a4paper]{svproc}

\usepackage{times}
\usepackage{epsfig}
\usepackage{graphicx}
\usepackage{amsmath}
\usepackage{amssymb}
\usepackage{algorithm}
\usepackage{algorithmic}

\usepackage{wrapfig}
\usepackage{xcolor}
\usepackage{url}
\usepackage{caption}
\usepackage{subcaption}

\long\def\invis#1{}
\usepackage{hyperref}
\usepackage{tikz}

\newcommand\copyrighttext{%
  \footnotesize \textcopyright 2023 Springer. Personal use of this material is permitted.
  Permission from Springer must be obtained for all other uses, in any current or future
  media, including reprinting/republishing this material for advertising or promotional
  purposes, creating new collective works, for resale or redistribution to servers or
  lists, or reuse of any copyrighted component of this work in other works.\\
  Accepted at the International Symposium on Experimental Robotics (ISER) 2023.%
  }
\newcommand\copyrightnotice{%
\begin{tikzpicture}[remember picture,overlay]
\node[anchor=south,yshift=20pt] at (current page.south) {\fbox{\parbox{\dimexpr\textwidth-\fboxsep-\fboxrule\relax}{\copyrighttext}}};
\end{tikzpicture}%
}

\let\subparagraph\paragraph
\usepackage[compact]{titlesec}

\titleformat{\paragraph}[runin]{\normalfont\normalsize\bfseries}{\theparagraph}{1em}{}
\titlespacing*{\paragraph}{0pt}{0.25ex plus 1ex minus .2ex}{1em}

\usepackage[style=nature,mincitenames=1,maxcitenames=2,maxbibnames=2,doi=false,url=false,isbn=false,date=year]{biblatex}

\begin{document}
\mainmatter              %
\title{Improving the perception of visual fiducial markers in the field using Adaptive Active Exposure Control}
\titlerunning{Adaptive active exposure control}
\author{Ziang Ren, Samuel Lensgraf, Alberto Quattrini Li}
\authorrunning{Ren et al.} %

\institute{Dartmouth College, Hanover NH 03755, USA,\\
\email{ziang.ren.gr@dartmouth.edu}\\}

\maketitle              %
\copyrightnotice
\begin{abstract}

Accurate localization is fundamental for autonomous underwater vehicles (AUVs) to carry out precise tasks, such as manipulation and construction. Vision-based solutions using fiducial marker are promising, but extremely challenging underwater because of harsh lighting condition underwater.

This paper introduces a gradient-based active camera exposure control method to tackle sharp lighting variations during image acquisition, which can establish better foundation for subsequent image enhancement procedures. Considering a typical scenario for underwater operations where visual tags are used, we proposed several experiments comparing our method with other state-of-the-art exposure control method including Active Exposure Control (AEC) and Gradient-based Exposure Control (GEC). Results show a significant improvement in the accuracy of robot localization. This method is an important component that can be used in visual-based state estimation pipeline to improve the overall localization accuracy. 

\end{abstract}

\section{Introduction}

\begin{wrapfigure}[11]{r}{0.35\textwidth}
  \vspace{-2.5em}
  \includegraphics[width=0.35\textwidth]{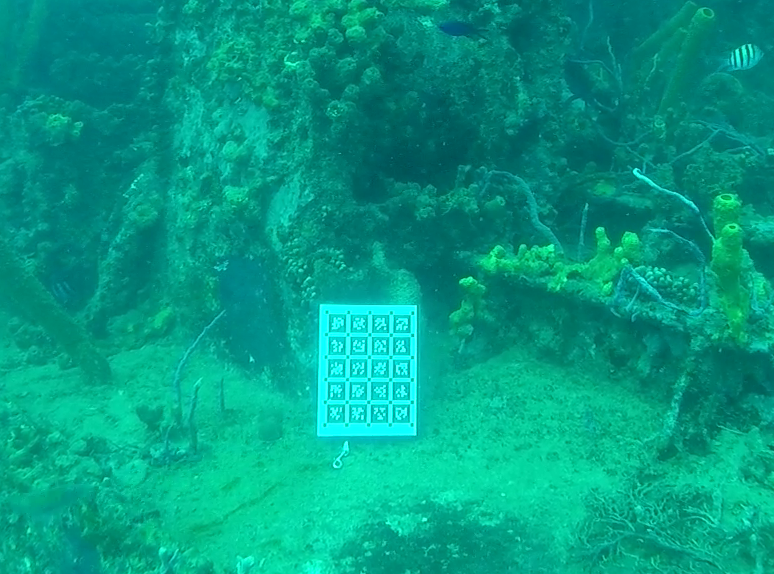}
  \vspace{-0.3in}
  \caption{Default exposure control over exposes visual fiducial.}
  \label{fig:fiducial-on-reef}
\end{wrapfigure}

We present a method for improving the sensing of relative position of visual fiducial markers in adversarial lighting scenarios commonly encountered in underwater deployments. Visual fiducial markers are commonly used to provide relative position information to panels~\cite{palomerasAutonomousIAUVDocking2014}, object transport~\cite{lensgrafBuoyancyEnabledAutonomous2023}, and docking~\cite{vivekanandanAutonomousUnderwaterDocking2023}. In marine domains, back-scatter and sharp lighting gradients make accurate sensing of visual fiducial markers challenging. Figure~\ref{fig:fiducial-on-reef} shows an example in the field in which the globally optimal exposure time is a poor choice for sensing a visual fiducial marker. We overcome these challenges by optimizing the exposure quality of the visual fiducial marker itself rather than over the image as a whole. By improving the stability and detection rate of visual fiducial markers, we can improve performance on downstream tasks.

Marine robotics is increasingly important in the face of increasing development of coastal areas for energy production. Localizing and stabilizing marine robots continue to be a challenge. One partial source of this problem is the lack of availability of reliable sensing information. In fact, a global positioning system, such as GPS, is not available. To overcome this challenge on the sensing side, we propose to more closely couple the control of the exposure time of the robot's on-board camera with the sensing of visual fiducial markers. Our approach can enable low-cost ways to provide high certainty reference points for marine robots in difficult environments.

\begin{wrapfigure}[16]{r}{0.35\textwidth}
    \vspace{-2.3em}
    \includegraphics[width=0.35\textwidth]{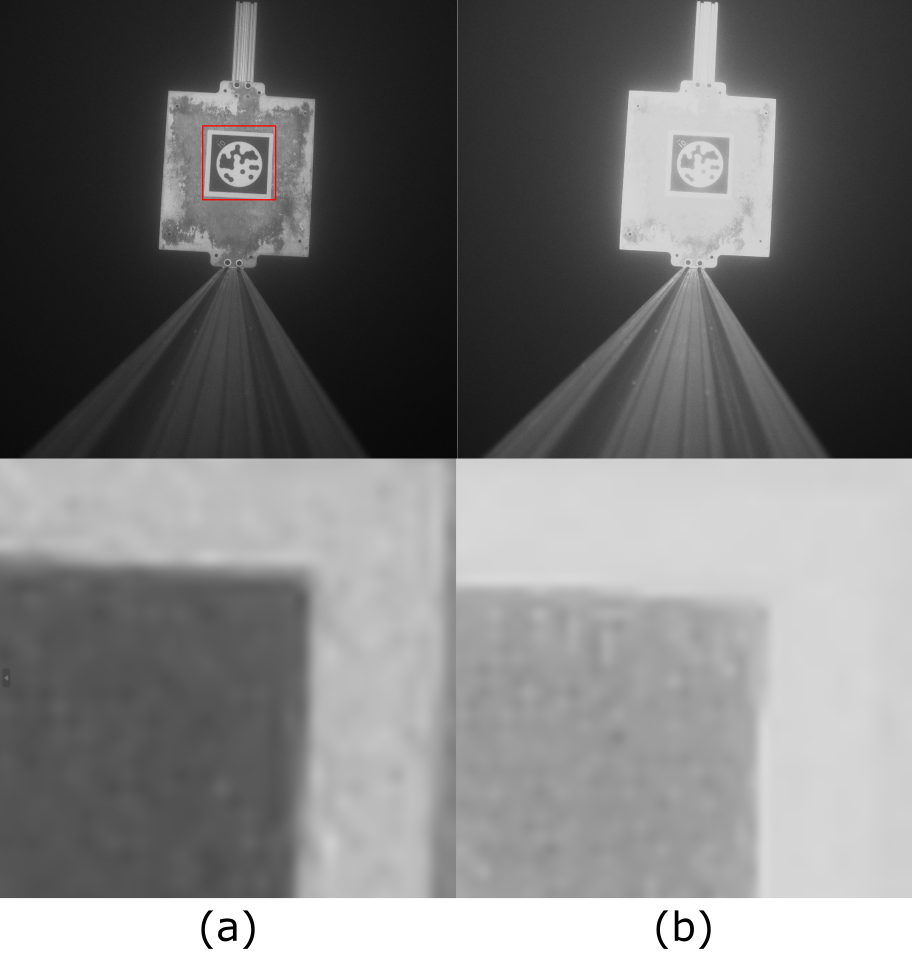}
    \vspace{-0.3in}
    \caption{Visual fiducial markers captured in the Connecticut River with global (b) and our proposed local (a) exposure optimization.}
    \label{fig:comparison-river-exposure-intro}
\end{wrapfigure}

The accuracy of visual fiducial marker sensing depends strongly on the sharpness of their corners in a captured image. Figure~\ref{fig:comparison-river-exposure-intro} compares the exposure of a visual fiducial marker achieved using our method against a global method. Our method, which we call \emph{adaptive active exposure control} (AAEC), improves the exposure quality of visual fiducial markers in lighting scenarios which commonly confuse global methods. 

We extend methods which have been tested for improving the sharpness of image for on-land robots~\cite{aec, aec2}. To adapt to the marine domain, we add a dynamic area of interest. The dynamic area of interest is re-computed in each image frame and used to track the desired object even when the robot moves. To make our algorithm more robust to rapid changes in lighting, we add momentum to the gradient descent used to optimize exposure time.

\section{Related work}

There are currently numerous methods which seek to improve the quality of images in underwater settings using post processing~\cite{roznereRealtimeModelbasedImage2019a, wangDeepCNNMethod2017a,islam2020fast,cho2018channel}. Post processing is an attractive method for improving images qualitatively, but they can provide no direct measure of actual improvement. Our approach can complement and extend post processing methods. By improving the base image quality, we can provide a basis for more performant post-processing algorithms, particularly when only small regions of an image are of interest. 

Meanwhile, image gradient holds crucial significance in feature detection. The EDPF algorithm \cite{EPDF} used by the specific fiducial marker employed in our experiment, STAG, also utilizes gradients
to detect edge segments: the task of joining the anchor points is done by
maximizing the total gradient response along its path \cite{benligiraySTagStableFiducial2017}. Therefore, we think it's highly worthwhile
to combine gradient information into the exposure control for detecting markers. Existing work on automatic exposure control utilizing gradient information focuses on improving the performance of SLAM and VO systems by optimizing images globally~\cite{aec,gec,wangAutomatedCameraexposureControl2022, beginAutoExposureAlgorithmEnhanced2022}. Our approach is the first specialized towards sensing visual fiducial markers in the field, and is the first to experimentally test the quality of visual fiducial sensing in underwater environments. The use of a specific region of interest makes the algorithm improve the sensing of visual markers and able to converge quickly.

Visual fiducial marker designs which provide high quality relative position sensing have been studied by many researchers~\cite{benligiraySTagStableFiducial2017, tianPolarTagInvisibleData2020, kalaitzakisFiducialMarkersPose2021, bergamascoRUNETagHighAccuracy2011}, however no existing approach combines visual fiducial marker sensing with active exposure control. By taking a more holistic approach to improving the exposure quality of visual fiducial markers, we can extend the utility of visual fiducials to realistic field environments.

\section{Technical approach}

\begin{wrapfigure}[10]{r}{0.4\textwidth}
  \vspace{-3em}
  \includegraphics[width=0.4\textwidth]{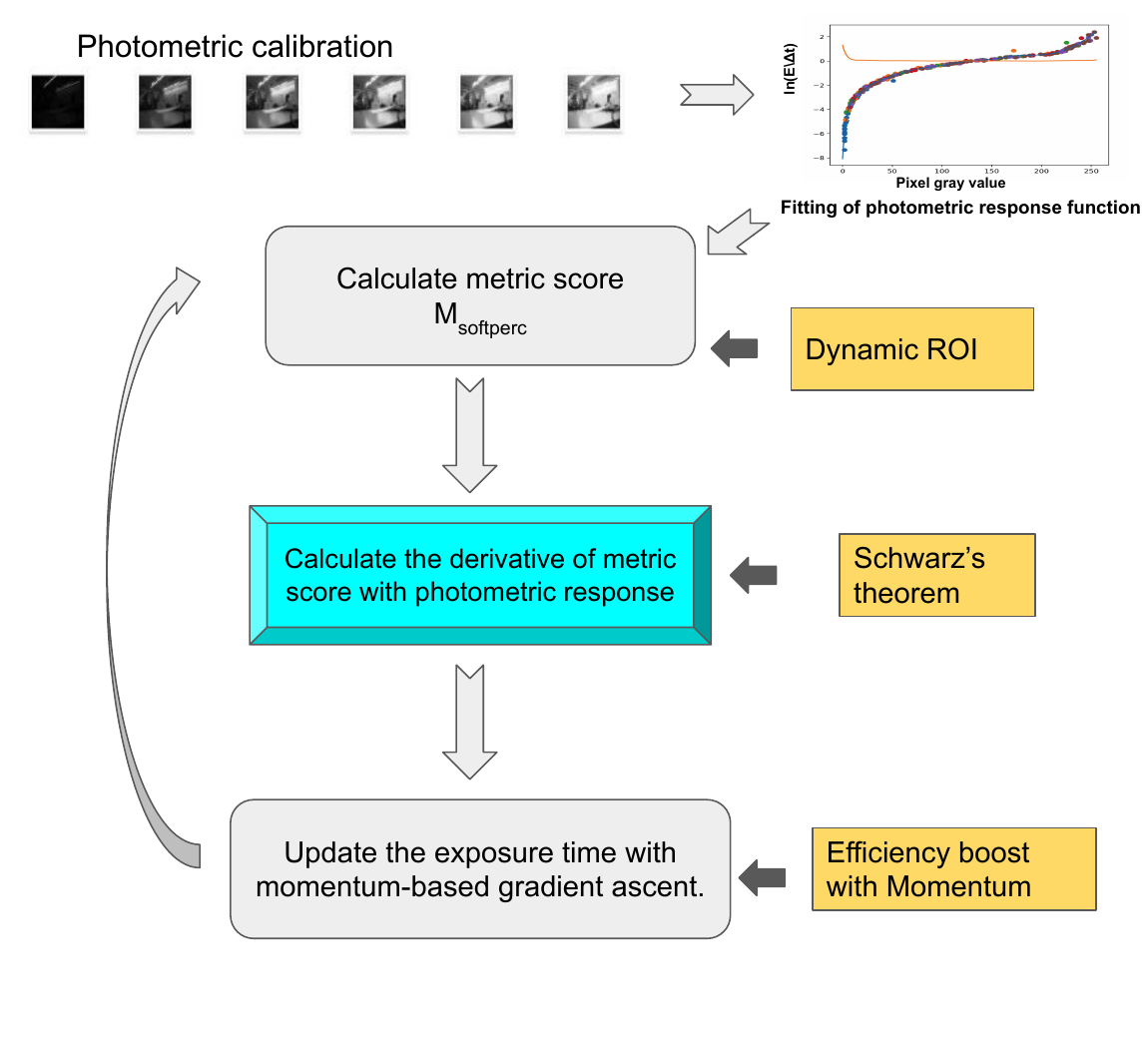}
  \vspace{-0.3in}
  \caption{The workflow of AAEC.}
  \label{fig:conv1}
\end{wrapfigure}

We optimize the detection of visual fiducial markers by adding dynamic exposure control into the detection pipeline. Our exposure control method computes gradient-based image quality metric over a small region of interest around the detected visual fiducial. 

\subsection{Method workflow}

Improved from the classic Active Exposure Control (AEC) proposed by \textcite{aec}, our AAEC (shown in Figure \ref{fig:conv1}) features subject awareness and a step-length self-adjusting gradient ascent. The principal idea can be summarized into two steps: first, to select an image quality metric containing gradient information. Second, update the exposure time frame-by-frame with gradient ascent. Algorithm \ref{algo1} details our approach.

\begin{algorithm} 
	\caption{Adaptive Active exposure control} 
	\label{alg:aec_algo} 
	\begin{algorithmic}[1]
            \WHILE{true}
	    \STATE Capture the image
            \STATE Detect Marker and derive RoI
            \STATE Compute Gradient within the RoI \textbf{G}
            \STATE Calculate metric score for the image within the RoI
            \STATE Calculate the derivative of the metric $\frac{\delta \Delta M_{\textit{softperc}}}{\delta \Delta t}$
	    \IF{$ \frac{\delta \Delta M_{\textit{softperc}}}{\delta \Delta t} \geq \textit{threshold}$}

        \STATE     $ \textit{Momentum}_{\textit{now}}= \gamma \textit{Momentum}_{\textit{prev}} + \iota \frac{\delta M_{\textit{softperc}}}{\delta \Delta t}$
          
	    \STATE     $\Delta t_{\textit{next}}= \Delta t +  \textit{Momentum}_{\textit{now}}$

	    \ENDIF
            \ENDWHILE
	\end{algorithmic} 
        \label{algo1}
\end{algorithm}

In Algorithm \ref{algo1}, ${M_{\textit{softperc}}}$ is an evaluation metric that contains the gradients information, which is also used by the original AEC proposed in \cite{aec}. It is calculated using the equation below:

\begin{equation}\label{eq:msoftperc}
\begin{aligned}
    M_{\textit{softperc}}(p)=\sum_{i \in [0,S]} W_{i} G_{i}
\end{aligned}
\end{equation}

\noindent where $W_{i}$ and $G_{i}$ are the weight and the gradient for the $i$-th pixel. The list $[0, S]$ indexes each pixel in the order of ascending gradient value. $W$ is defined to emphasize a certain range of gradients. In practice, we set $W$ to emphasize gradients based on a desired percentile range. We found that a emphasizing percentile range of 0.6 to 0.9 produces good results. We compute the weight $W$ using Equation~\ref{eqn:weight-metric}.

\begin{equation}
\begin{aligned}
    W_{i}=\left\{
\begin{aligned}
\frac{1}{N} \sin(\frac{\pi}{2\lfloor pS \rfloor } i)^k & , & i \leq \lfloor pS \rfloor, \\
\frac{1}{N} \sin(\frac{\pi}{2}-\frac{\pi (i-\lfloor pS \rfloor)}{2(S-\lfloor pS \rfloor) } i)^k & , & i > \lfloor pS \rfloor,
\end{aligned}
\right.
\end{aligned}
\label{eqn:weight-metric}
\end{equation}

$W_i$ is continuous and has a sharp peak near the desired percentile, $p$, of pixel gradients. It is assumed that the range $[0,S]$ is sorted by the gradient intensity.

The calculation of $\frac{\delta \Delta M_{\textit{softperc}}}{\delta \Delta t}$ requires the connection between gradient $G$ and the exposure time $\Delta t$:

\begin{equation}
\begin{aligned}
    \frac{\delta M_{\textit{softperc}}}{\delta \Delta t}=\sum_{i\in[0,S]}
    {W_{i}\frac{\delta G_{i}}{\delta \Delta t}}
\end{aligned}
\end{equation}

\noindent Computing $\frac{\delta G_{i}}{\delta \Delta t}$ requires applying differentiation rules and we refer the reader to~\cite{aec} for the full derivation.

\subsection{Primary features and gains}

\subsubsection{Momentum-based self-adjusting gradient ascent}

In order to apply our work to a moving robot, it's critical to achieve fast convergence to an optimal exposure time. Therefore, we introduce the concept of momentum into our optimization of exposure time to greatly reduce the number of frames required for the exposure time to converge. Like a ball rolling down hill, the continuous descent increases the ball's velocity, enabling it to reach the bottom faster~\cite{momentum}. In our case, momentum and its relationship to our gradient ascent is mathematically represented as:
\begin{equation}
\begin{aligned}
\label{eq1}
& v_t=\gamma  v_{t-1} + \eta M’_{\textit{softperc}}(\Delta T_{t}) \\
& \Delta T_{t+1}=\Delta T_{t} + v_t
\end{aligned}
\end{equation}
where $ v_t$ is the momentum for the current frame $t$, $\gamma$ is the factor controlling the weight of previous accumulated momentum, $\eta $ is the learning rate, $\Delta T_{t}$ is the exposure time for frame $t$, $ M’_{\textit{softperc}}(\Delta T_{t})$ is the first derivative of $M_{\textit{softperc}}$ with respect to exposure time. While a higher $ M_{\textit{softperc}}$ represents a better image, the momentum will be accumulated if $ M_{\textit{softperc}}$ is continuously increasing, leading to a faster convergence. Our tests suggest that this increases the speed by 560\%  under normal lighting conditions.

To address potential issues brought by the gradient of unwanted image noise, the range for gradient information of interest needs to be carefully defined. Therefore, leveraging the heavily-tailed distribution of image gradients~\cite{heavy1, heavy2}, $M_{\textit{softperc}}$, shown in Eq.~\eqref{eq:msoftperc},  %
is applied in the gradient ascent process.

\subsubsection{Subject awareness}

Harsh changes in lighting conditions, which is commonly encountered underwater, will lead to unstable marker acquisition if exposure control is performed without subject awareness. Additionally, the original AEC\invis{ and GEC are} is slow when processing high-resolution images. This could reduce the convergence speed, making downstream tasks more difficult to achieve. We use a region of interest (RoI) that is dynamically computed around the fiducial marker to solve these problems, which drastically increases the processing speed and corresponding accuracy. In determining the RoI, our approach commences by identifying the four corners of the detected markers. The upper left corner of the preliminary bounding box is determined by the corners' minimal x and y coordinates, while the lower right is ascertained by their maximum counterparts. To ensure the encompassed region is slightly more expansive than the marker itself, providing additional context, a padding of 10\% derived from the width and height of this initial bounding box is applied. This augmented area is thus designated as our RoI.

\section{Experiments}

\begin{figure}[!b]
     \centering
     \begin{subfigure}[b]{0.45\textwidth}
         \centering
         \includegraphics[width=\textwidth]{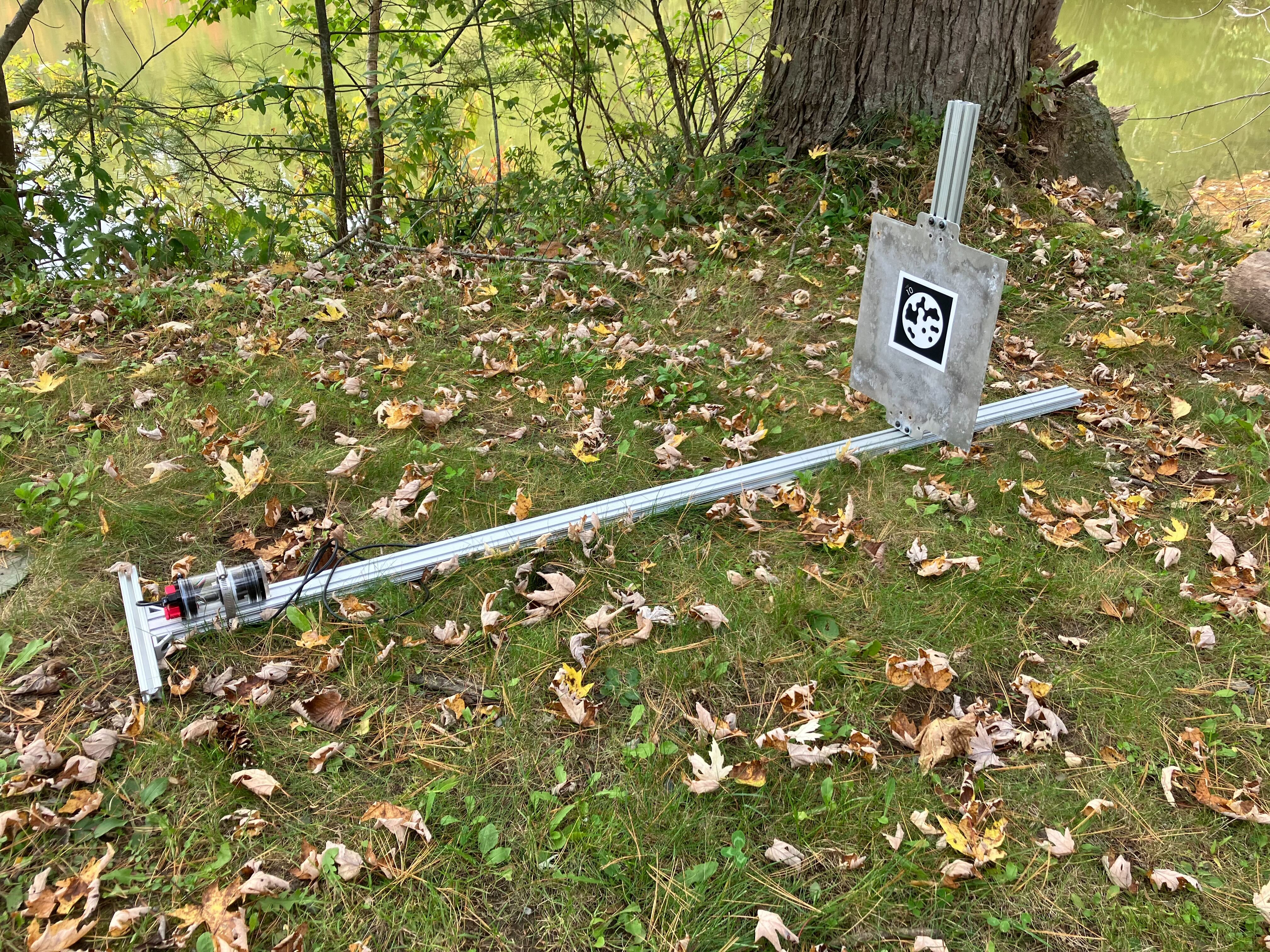}
         \caption{}
     \end{subfigure}
     \begin{subfigure}[b]{0.45\textwidth}
         \centering
         \includegraphics[width=\textwidth]{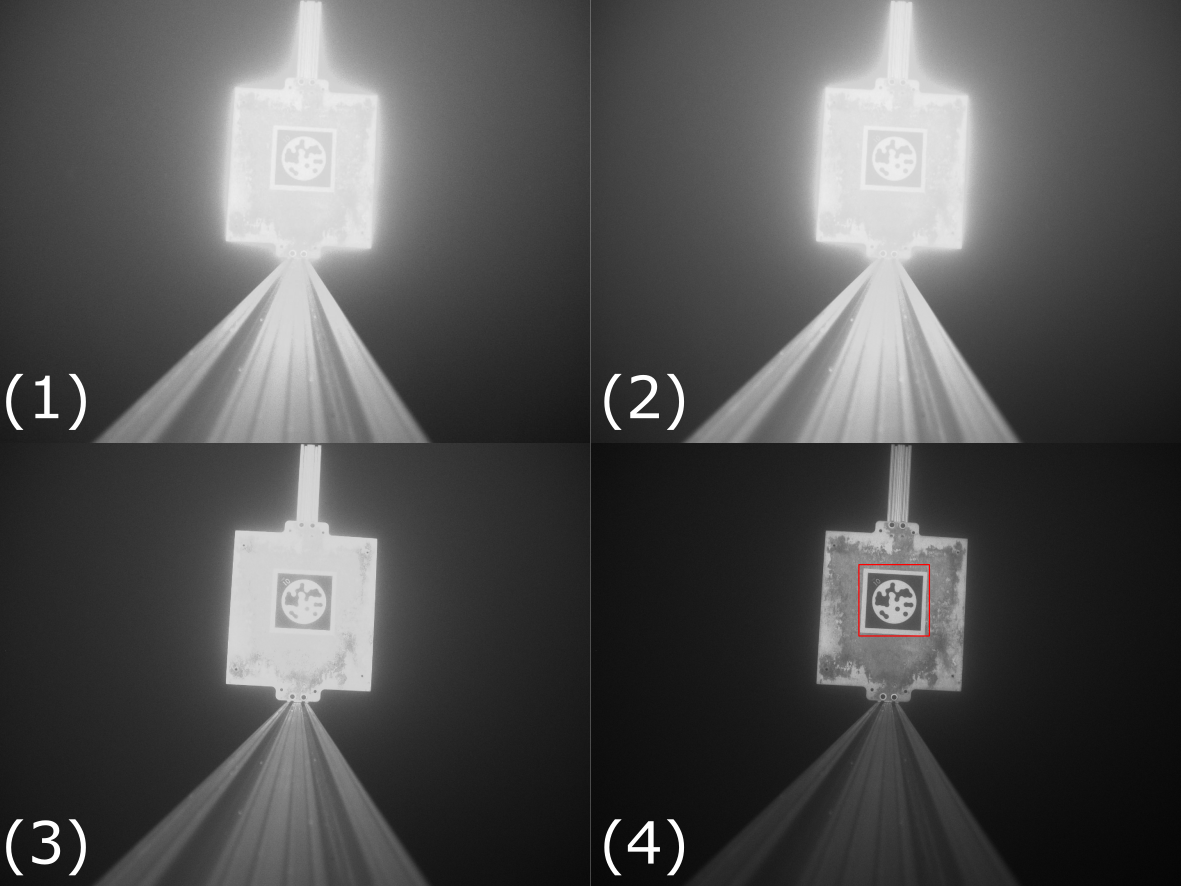}
         \caption{}
     \end{subfigure}
        \caption{(a) Testing rig used for field testing our approach in the Connecticut river. (b) Comparison of algorithms (1) default, (2) AEC, (3) GEC, (4) AAEC in river water.}
        \label{fig:river-results-algorithm-comparison}
\end{figure}

To measure the effectiveness of our approach, we compare our exposure control method against three other approaches. As a baseline, we compare against the built-in exposure control method used in our camera (FLIR Blackfly S~\cite{BlackflyUSB3Teledyne}). We call this baseline~\textit{default} in the following discussion. We also implemented two state-of-the-art global exposure control algorithms from the literature. Gradient-based exposure control, GEC~\cite{gec}, was developed to improve image quality for computer vision applications on mobile robots. We also implemented the global active exposure control method reported by Zhang et al.~\cite{aec}, which we call AEC. The AEC algorithm has been shown to improve the performance for visual odometry tasks. Our specialized implementation, is referred to as AAEC.

We evaluate the precision of the detected relative position using the determinant of the covariance matrix of the computed relative positions. The determinant of the covariance matrix is the multiplication of its three eigenvalues. Each eigenvalue represents the variance along one of the principal axes of an ellipsoid. Multiplying the three eigenvalues gives us a sense of the overall precision. Higher covariance determinants signify more uncertainty. We also evaluate the detection rate and distance to a ground truth point of reference where it is relevant to give a concrete idea of the accuracy of measured positions. 

\subsection{Field experiment: Connecticut River}
\begin{wrapfigure}[13]{r}{0.4\textwidth}
  \vspace{-1em}
  \centering
   \includegraphics[width=0.4\textwidth,trim={2cm 1cm 2cm 1cm},clip]{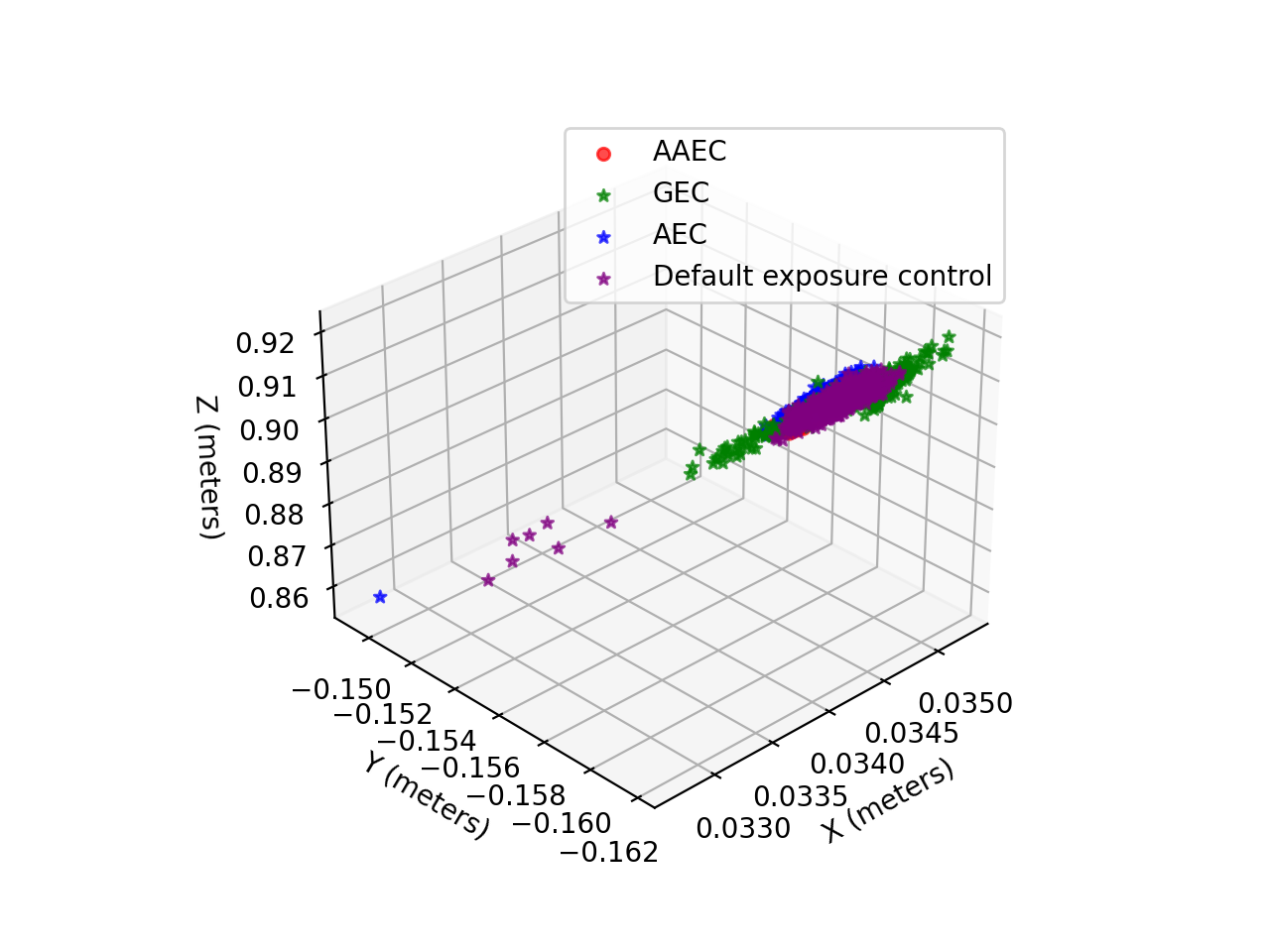}
  \caption{Measured positions for different exposure control methods for river field test.}
\label{fig:scatter-plot-river-experiment}
\end{wrapfigure}
To evaluate the effectiveness of our exposure control algorithm in a real-world aquatic setting, we conducted extensive field tests in the Connecticut River. Our experimental setup comprised a camera securely mounted on one end of a specialized rig. Positioned in front of this camera, on the opposite end of the rig, was the fiducial marker. With this arrangement, we submerged the entire assembly underwater, ensuring that the camera's field of view was consistently directed towards the marker. Over the duration of each test, which lasted one minute, the system continuously recorded the detected positions of the fiducial marker, enabling us to gauge the algorithm's precision and reliability in real underwater conditions. 

Figure~\ref{fig:river-results-algorithm-comparison} shows our testing rig on land and the image quality after the automatic exposure control converged. Our testing rig deployed in the river simulates a particularly challenging underwater imaging scenario. In dark or murky water, a robot will often be presented with cases where its head lamps illuminate a small region requiring detailed sensing while much of the image is dark.

\begin{table}[h]
\vspace{0.0cm}
\renewcommand{\arraystretch}{1.0}
\centering
\vspace{-0.0em}
\scalebox{0.95}{
\begin{tabular}{cccccc}
\hline
      & AAEC & GEC & AEC & Default\\\hline
Determinant & $\mathbf{1.65\times10^{-23}}$ & $1.02\times10^{-20}$ & $1.38\times10^{-20}$ & $3.82\times10^{-20}$ \\
Detection rate & 100\% & 100\% & 100\% & 99.60\% \\
Maximum Distance (meters) & $\mathbf{0.0040}$ & $0.0273$ & $0.0598$ & $0.0462$ \\
\hline
\vspace{0.1em}
\end{tabular}
}
\caption{Determinant of covariance matrix, marker detection rate, and maximum distance (in meters) between detected locations for 4 exposure control methods. Smaller determinant represents more concentrated acquired marker locations (lower is better).}
\label{tab:river-result-table}
\vspace{-2.0em}
\end{table}

Table~\ref{tab:river-result-table} shows the results of our testing in the Connecticut river. Our AAEC algorithm outperforms the other algorithms by three orders of magnitude in terms of the precision of the set of the measured positions. While each algorithm was able to reliably detect the visual fiducial marker, ours achieved both high precision and did not have outliers which could confuse control algorithms. Figure~\ref{fig:scatter-plot-river-experiment} shows a scatter plot of the sensed positions for the fixed camera, in which the positions detected by our algorithm are subsumed by the point clouds representing the other algorithms.

\subsection{Motion tracking}

\begin{figure}[!t]
    \centering
    \includegraphics[width=0.8\textwidth]{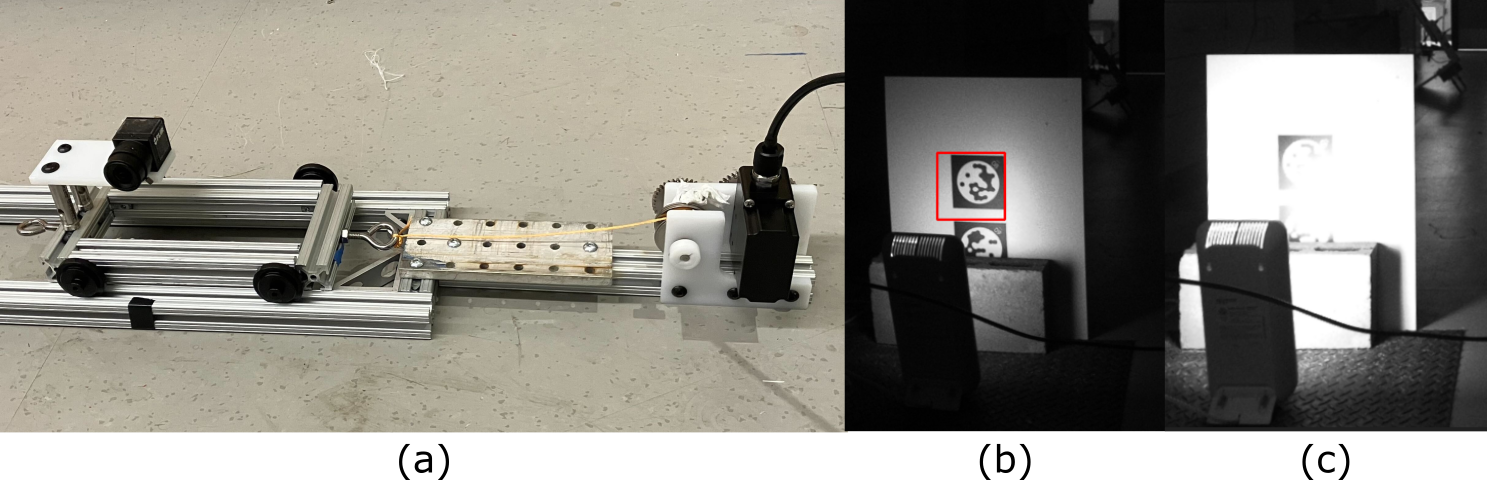}
    \caption{(a) Servo-based rig for testing motion tracking. View of visual fiducial under adversarial lighting with (b) AAEC and (c) built in exposure control.}
    \label{fig:camera-motion-testing-rig}
\end{figure}

To test the ability of our approach to track moving markers, we built a testing rig which uses a servo motor to move a cart bearing a camera. Figure~\ref{fig:camera-motion-testing-rig} shows the rig used to test camera motions. A continuous rotation servo (can be seen in the right part of Figure~\ref{fig:camera-motion-testing-rig} (a)) winds a string used to move a cart bearing the camera laterally at a fixed speed.

We compare the detection rate and accuracy of the tracked motion for each of the four algorithms in three different lighting conditions. We tested flat overhead lighting as a baseline, which we refer to as default or normal lighting. Second, we tested a low light setting, common in marine domains, and finally, we tested an adversarial lighting condition. For the adversarial lighting condition, we used a bright fore-light on the marker and a dark background. This scenario simulates a similar scenario to the one shown in Figure~\ref{fig:river-results-algorithm-comparison}: when a robot approaches an object to sense in an otherwise feature-sparse area.

\begin{table}[!t]
\centering
\begin{minipage}{.47\linewidth}
\renewcommand{\arraystretch}{1.0}
\centering
\scalebox{0.75}{
\begin{tabular}{ccccc}
\hline
      & {Adversarial lighting }  & { Normal lighting } & {Low light } &\\\hline
			{AAEC}& {$\mathbf{0.0058}$}&{$0.0076$}&          {$0.0066$}& \\
                {GEC}& {$0.0490$}& {$0.0093$}& {\textbf{$0.0094$}}&\\
                {AEC}&{No detection}& {$\mathbf{0.0075}$}&{$\mathbf{0.0064}$}&\\
    		{Default}&{No detection}&{$0.0092$}&            {\textbf{$0.0072$}}&\\
\hline
\end{tabular}
}

\caption{Average shortest distance from the detected locations to the ground truth of moving trajectory (meters).}
\label{tab:motion-tracking-results}
\end{minipage}%
~~\begin{minipage}{.47\linewidth}
\renewcommand{\arraystretch}{1.0}
\centering
\scalebox{0.75}{
\begin{tabular}{ccccc}
\hline
      & {Adversarial lighting }  & { Normal lighting } & {Low light} &\\\hline
        {AAEC}& {$\mathbf{99.32\%}$}& {$98.57\%$}&{$\mathbf{100.00\%}$}& \\
        {GEC}& {$29.29\%$}& {$96.82\%$}&{\textbf{$99.36\%$}}&  \\
        {AEC}& {$0\%$}& {$\mathbf{99.11\%}$}&
        {$94.85\%$}&\\
        {Default} & {$0\%$}& {$96.82\%$}   &{\textbf{$99.31\%$}}&\\
\hline
\end{tabular}
}
\caption{Marker detection rate for 4 exposure methods in motion tracking experiment.}
\label{tab:rate-motion}
\end{minipage}%
\vspace{-3em}
\end{table}

\begin{figure}[!t]
    \centering
    \includegraphics[width=.9\textwidth]{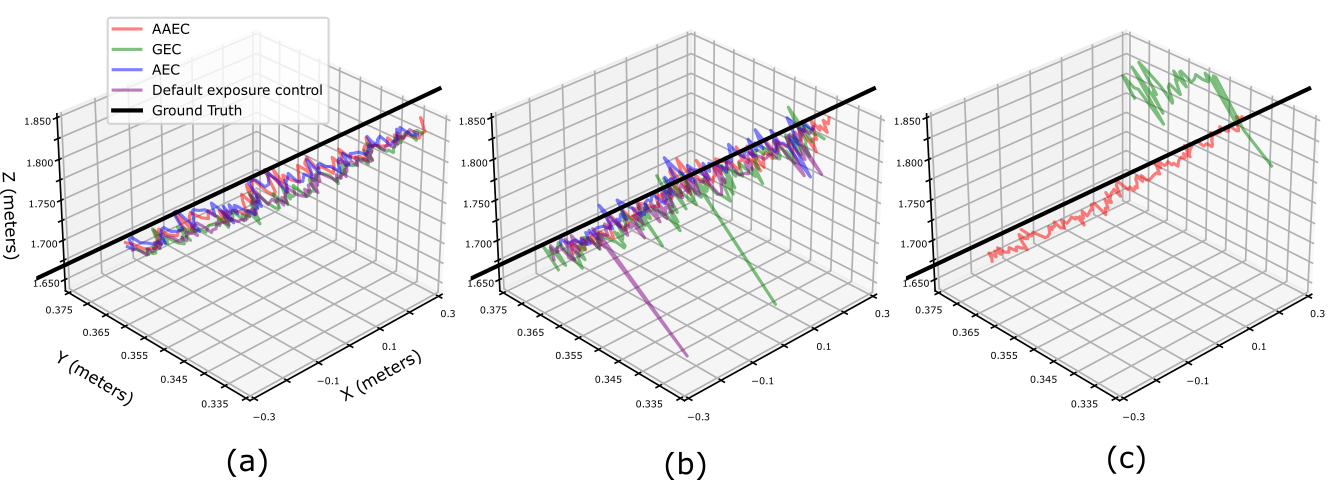}
    \caption{Recorded positions while moving the camera along a horizontal plane in (a) normal lighting, (b) low light, (c) adversarial front lighting. The black line marks measured ground truth.}
    \label{fig:scatter-plots-motion-experiment}
\end{figure}

Table~\ref{tab:motion-tracking-results} and Table~\ref{tab:rate-motion}  show the results of our motion tracking experiments. In the two non-adversarial lighting scenarios, the AAEC algorithm performs on par with the other baseline algorithms. However, in the adversarial scenario, AAEC outperforms all others by an order of magnitude in terms of the distance between the measured positions and the estimated trajectory. In the adversarial lighting scenario, the default and global AEC were unable to track the marker at all. This was due to a bright glare washing out a large portion of the visual fiducial marker. The measured and estimated trajectories for the tested algorithms are shown in Figure~\ref{fig:scatter-plots-motion-experiment}.

\subsection{Convergence time}

Our approach includes two additions to speed convergence: momentum and a small region of interest. To measure their impact, we perform an ablation study where we measure the time to convergence in three lighting scenarios: adversarial lighting, normal lighting, and  low light. For each lighting condition, we ensure that the initial step length for gradient ascent is the same for each exposure method. 

In Table \ref{tab:cov-time}, we observe a reduction in convergence time when the RoI is introduced in some scenarios such as adversarial lighting and low-light conditions. This reduction can be attributed to the fact that by narrowing down to a specific RoI, the system focuses on a smaller set of significant data, thereby optimizing its performance and reaching convergence faster.

\begin{table}[h]
\vspace{-1em}
\renewcommand{\arraystretch}{1.0}
\centering
\scalebox{0.9}{
\begin{tabular}{ccccc}
\hline
      & {Adversarial lighting }  & { Normal lighting } & {Low light} &\\\hline
        {AAEC (Momemtum+RoI)}& {$\mathbf{4}$}& {$\mathbf{3}$}&{$\mathbf{6}$}& \\
        {AAEC (RoI only)}& {$23$}& {$42$}&{\textbf{$43$}}&  \\
        {AEC}& {$63$}& {$43$}&
        {$72$}&\\
\hline
\end{tabular}
}
\caption{Comparison of convergence time cost (in seconds) for exposure.}
\label{tab:cov-time}
\vspace{-3em}
\end{table}

Meanwhile, the addition of momentum further enhances this optimization. By integrating the previous weight update, momentum ensures smoother convergence and deters oscillations. With momentum, the time cost across all scenarios sees a significant decrease.

\subsection{Station keeping AUV}

\begin{wrapfigure}[16]{r}{0.34\textwidth}
    \vspace{-2.5em}
    \includegraphics[width=0.34\textwidth]{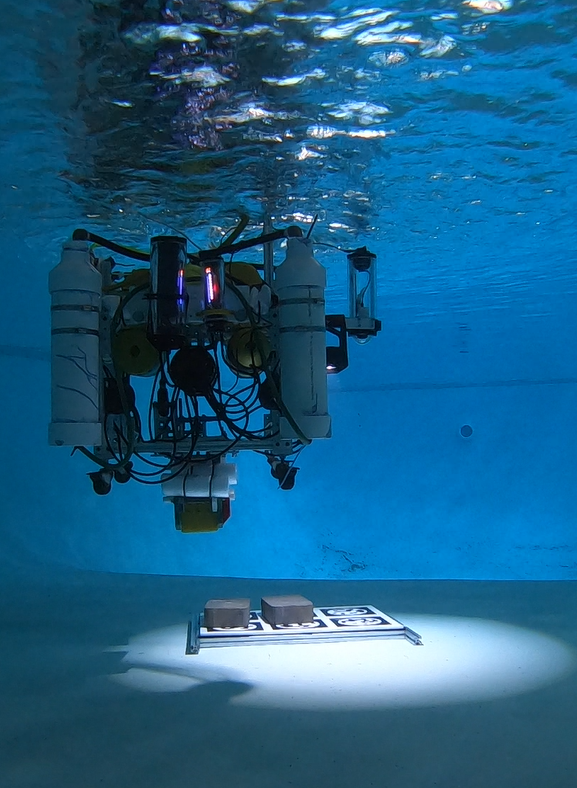}
    \vspace{-2em}
    \caption{AUV station keeping in low light test.}
    \label{fig:station-keeping-auv}
\end{wrapfigure}

Finally, to test our algorithm in tandem with a robot moving under its own control, we tasked our autonomous underwater vehicle (AUV) with station keeping using the results from a visual fiducial marker. The AUV is controlled using PID controller in six degrees of freedom. For each test, we started the AUV near the target location and initialized the controller. The AUV uses real time position feedback from a marker viewed through a fisheye lens to position itself. Figure~\ref{fig:station-keeping-auv} shows our AUV system station keeping during the low light test. After an initial period in which the AUV's controllers are allowed to stabilize, we record the sensed relative positions to the visual fiducial marker.

Tables~\ref{tab:station-keeping-results} and~\ref{tab:station-keeping-detection-rate} show the results of our station keeping test. In this test, noise from the AUV's control system (i.e., overshoot from the controller) is combined with noise from the sensed position of the visual fiducial markers. The noise of the sensed marker positions in the low and normal lighting scenarios is improved by our AAEC algorithm. This shows that even with the relatively unpredictable motions a free-floating robot is subject to, our region of interest can be dynamically updated and the exposure quality maintained. In the adversarial lighting scenario, our method out performs the others in terms of both marker detection rate and position sensing precision. 

\begin{table}[!t]
\centering
\begin{minipage}{.47\linewidth}
\renewcommand{\arraystretch}{1.0}
\centering
\scalebox{0.73}{
\begin{tabular}{ccccc}
\hline
      & {Adversarial lighting }  & { Normal lighting } & {Low light} &\\\hline
        {AAEC}& {$\mathbf{8.65\times10^{-13}}$}& {$\mathbf{9.79\times10^{-13}}$}&{$\mathbf{2.81\times10^{-13}}$}& \\
        {GEC}& {$8.74\times10^{-12}$}& {$3.62\times10^{-12}$}&{$1.36\times10^{-12}$}&  \\
        {AEC}& {$2.45\times10^{-12}$}& {$2.25\times10^{-12}$}&
        {$1.05\times10^{-12}$}&\\
        {Default} & {$4.73\times10^{-12}$}& {$3.92\times10^{-12}$}   &{$5.20\times10^{-13}$}&\\
\hline
\end{tabular}
}
\caption{Determinant of Covariance matrix.}
\label{tab:station-keeping-results}
\end{minipage}~~~
\begin{minipage}{.47\linewidth}
\renewcommand{\arraystretch}{1.0}
\centering
\scalebox{0.73}{
\begin{tabular}{ccccc}
\hline
      & {Adversarial lighting }  & { Normal lighting } & {Low light} &\\\hline
        {AAEC}& {$\mathbf{67.79\%}$}& {$67.25\%$}&{$58.08\%$}& \\
        {GEC}& {$51.46\%$}& {$\mathbf{73.51\%}$}&{$59.24\%$}&  \\
        {AEC}& {$61.42\%$}& {$53.80\%$}&
        {$57.51\%$}&\\
        {Default} & {$44.33\%$}& {$46.65\%$}   &{$\mathbf{59.76\%}$}&\\
\hline
\end{tabular}
}
\caption{Detection rate for station keeping experiment.}
\label{tab:station-keeping-detection-rate}
\end{minipage}
\vspace{-3em}
\end{table}

\section{Conclusion \& future work}

In this paper, we have presented the first study of using active and adaptive exposure control for improving the sensing of visual fiducial markers in the field. Our work shows that by properly tuning the exposure time used when sensing visual fiducial markers, we can improve sensing performance significantly. This leads to the idea that by considering the camera and visual fiducial markers together as a unit which work together to localize the robot, we can improve the results robots can achieve in real world settings.

In our future work, we aim to adapt our approach for multi-target scenarios by incorporating multiple regions of interest (RoIs). By doing so, our exposure quality metric can guide higher-level RoI algorithms on prioritizing specific regions for data collection. Furthermore, we are examining the potential benefits and challenges of using down-sampled images, especially when paired with multiple RoIs, given some initial instability observed in our testing on using down-sampled images with single RoI.

In rare cases, our algorithm can fail to converge to a stable exposure time even for a small region of interest. We've observed this in cases where caustics dominate the region of interest. In these cases, the initial step length which works for most lighting scenarios causes oscillation in the final converged result. This can be solved by manually designating a smaller step length during the configuration phase. However, to maximize the automation of the system, we plan to explore dynamic step length assignment strategies which take into account the image quality in the region of interest. Meanwhile, it would be beneficial to develop the mechanism that use controlled light pulse to simulate caustics for experiment and testing. 

Finally, we plan to integrate our method together with a visual odometry package specialized towards operating on visual fiducial markers. By combining the absolute position sensing capabilities of visual fiducial markers, high quality imaging, and stable velocity estimated, we can create visual fiducials that are robust in the field and provide high rate information for control.

\section*{Acknowledgment}

This work was supported in part by NSF CNS-1919647, OIA-2024541, 2144624. Special thanks to Mingi Jeong, Luyang Zhao and Yitao Jiang for their supports. 

\printbibliography

\end{document}